%% file: main.tex
\documentclass[10pt,twocolumn,letterpaper]{article}

\usepackage{cvpr}              


\definecolor{cvprblue}{rgb}{0.21,0.49,0.74}
\usepackage[pagebackref,breaklinks,colorlinks,allcolors=cvprblue]{hyperref}

\usepackage{amsmath}

\def\paperID{19} 
\def\confName{CVPR}
\def\confYear{2025}

\title{EV-Flying: an Event-based Dataset for In-The-Wild Recognition of Flying Objects}

\author{Gabriele Magrini\\
University of Florence\\
{\tt\small gabriele.magrini@unifi.it}
\and
Federico Becattini\\
University of Siena\\
{\tt\small federico.becattini@unisi.it}
\and
Giovanni Colombo\\
University of Florence\\
{\tt\small giovanni.colombo@edu.unifi.it}
\and
Pietro Pala\\
University of Florence\\
{\tt\small pietro.pala@unifi.it}
}
\newcommand{\dataset}{EV-Flying}

\begin{document}
\maketitle

\begin{abstract}
Monitoring aerial objects is crucial for security, wildlife conservation, and environmental studies. Traditional RGB-based approaches struggle with challenges such as scale variations, motion blur, and high-speed object movements, especially for small flying entities like insects and drones. In this work, we explore the potential of event-based vision for detecting and recognizing flying objects, in particular animals that may not follow short and long-term predictable patters. Event cameras offer high temporal resolution, low latency, and robustness to motion blur, making them well-suited for this task. We introduce \dataset{}, an event-based dataset of flying objects, comprising manually annotated birds, insects and drones with spatio-temporal bounding boxes and track identities. To effectively process the asynchronous event streams, we employ a point-based approach leveraging lightweight architectures inspired by PointNet. Our study investigates the classification  of flying objects using point cloud-based event representations. The proposed dataset and methodology pave the way for more efficient and reliable aerial object recognition in real-world scenarios.\end{abstract}

\section{Introduction}
Monitoring flying objects in the sky has gained interest from several perspectives. On the one hand, monitoring the sky for aerial objects is a matter of primary importance for security, as flying devices such as drones are more and more diffused. Even toy drones, when misused, can pose privacy issues as they are often equipped with HD cameras or can enter, willingly or by mistake, into restricted fly zones such as airports.
On the other hand, many other kinds of flying objects exist. Birds, as well as insects, can be found in large quantities when observing the sky. Being able to recognize them is an equally important challenge as big flocks or even individual birds of prey can represent a danger for manned or unmanned aerial devices. 
The applications are not limited to security and safety. Environmental monitoring is another important facet of flying object recognition. Automatic detection and recognition of birds and insects is an open challenge in biology, where camera traps are often use to study their behaviors \cite{o2008picture}.
In addition, flying animals may also be easily confused with other objects of interest that one may want to detect (e.g., drones).

\begin{figure}[t]
    \centering
    \includegraphics[width=\columnwidth]{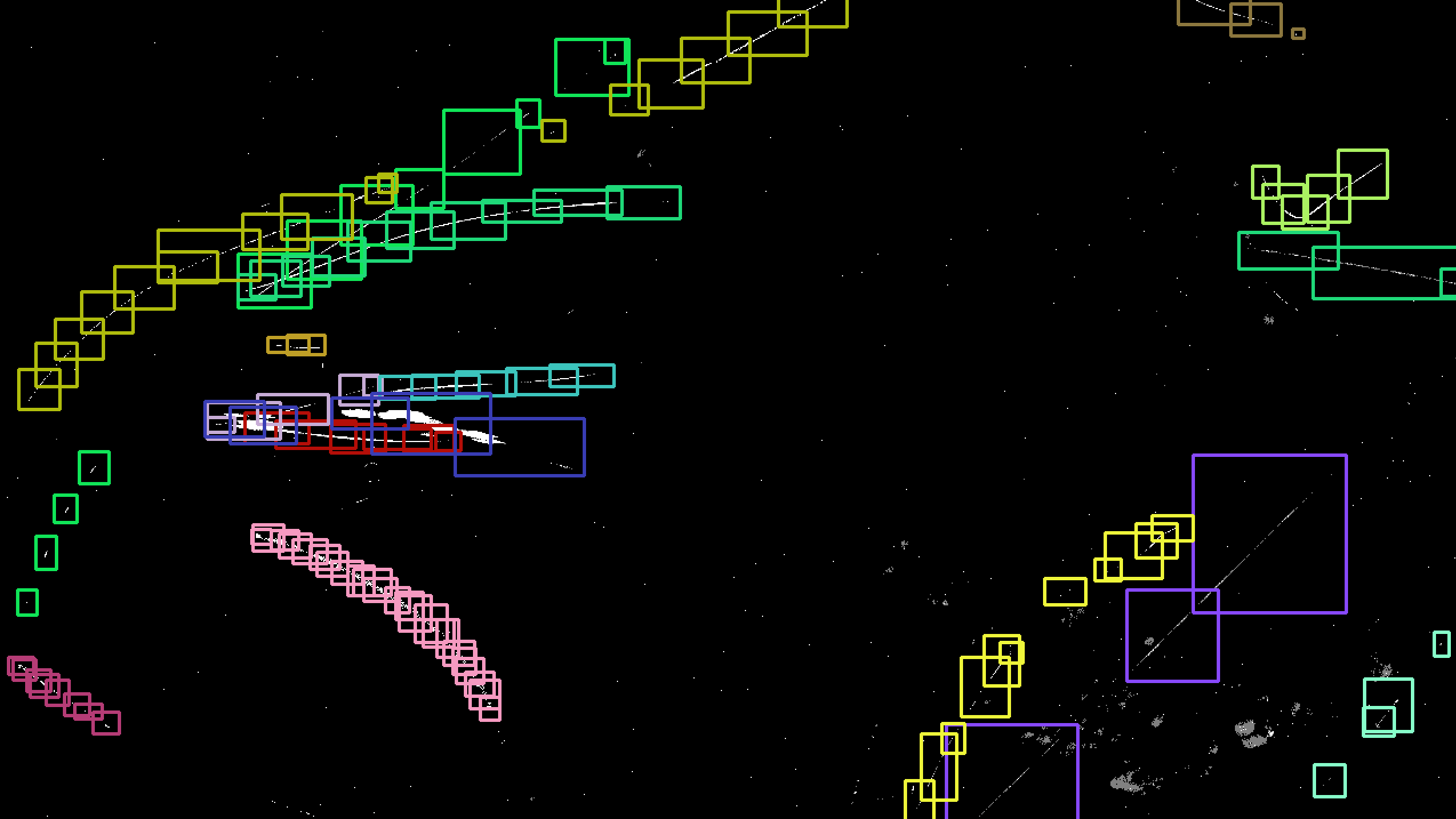}
    \caption{We propose \dataset{}, the first dataset of event-based flying objects, including birds, insects and drones. We manually annotate each object with bounding boxes at 30Hz and track identities.}
    \label{fig:eyecatcher}
\end{figure}

In general, recognizing different flying objects is a hard task. The large variability in scale, pose, orientation and distance poses at the same time two opposite challenges: first, the large diversity of appearances makes it necessary to learn a large array of visual and motion patterns; second, scale ambiguity makes it very easy to confuse small nearby objects with larger far away objects (an insect close to the camera may appear as a far away airplane).

Scale in particular is an issue that must be addressed. Whereas detecting macroscopic objects (e.g., planes, helicopters) is hard by itself, detecting small or far away flying objects poses some challenges that are not easy to address with standard cameras. Far away objects can be recognized with specific optics with long focal lengths, yet tiny objects as insects exhibit a low contrast with the background and, since they move very quickly compared to their dimension, they can generate a motion blur that impedes accurate recognition.
High-framerate cameras have been used to address these issues \cite{bjerge2022real, vo2024high}, although the computational burden of processing a large amount of frames per second if often too high for low-powered devices such as camera traps.

Recently, an effective alternative to standard RGB cameras has been proposed thanks to the usage of neuromorphic sensors \cite{gallego2020event}. Such devices, commonly known as event cameras, implement a paradigm shift from traditional vision sensors, moving from synchronous frames to asynchronous streams of events, i.e. pixel-wise illumination changes. The advantage of event cameras is that they generate a signal only when an illumination change is detected, making them power-efficient. Interestingly, such illumination changes can be captured at a micro-second rate, i.e. capturing extremely fine motion patterns. As a consequence, the generated event streams have negligible motion blur, making them suitable for detecting and tracking extremely fast objects. Since an illumination log-scale is used, neuromorphic sensors exhibit a high dynamic range, working also under scarce illumination conditions.

A few works have been proposed in recent literature, implementing a neuromorphic camera trap for monitoring wildlife, including medium-sized mammals \cite{hamann2024low} and insects \cite{gebauer2024towards}.
Event cameras have also proven effective for drone detection \cite{magrini2024neuromorphic}, offering large improvements against RGB counterparts.
Yet a comprehensive study of flying objects from a neuromorphic perspective still poses an open challenge.
Motivated by these findings, in this work we propose an event-based approach for detecting and recognizing flying objects.
We follow a point-based approach, as discriminative motion patterns, such as rotating drone propellers and flapping insect wings, can be extremely fast. To maintain temporally high recognition rates, we employ lightweight architectures based on the popular PointNet \cite{qi2017pointnet} to extract features directly from event clouds.

\begin{figure}
    \includegraphics[width=\linewidth]{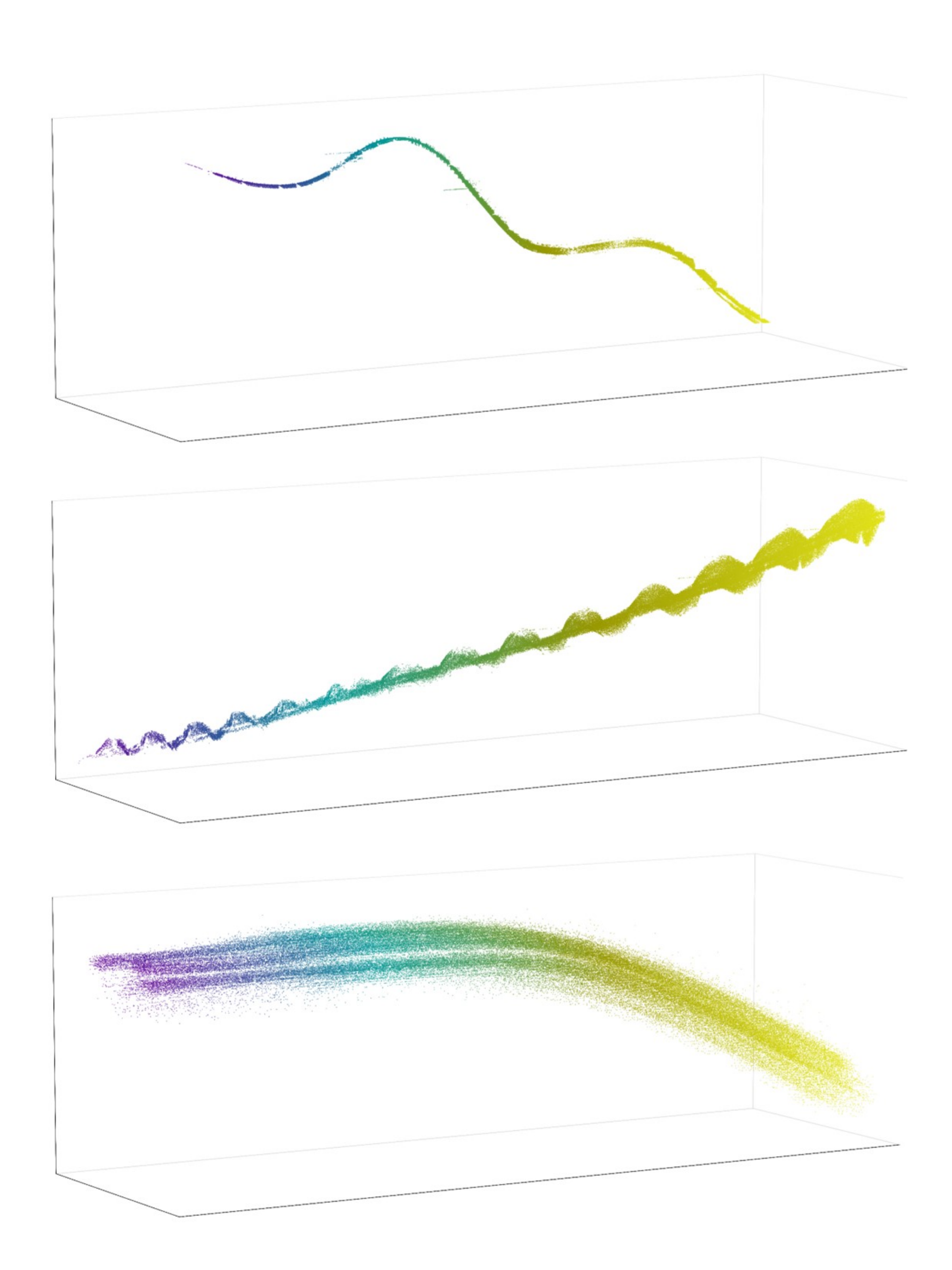}
    \caption{Samples of object tracks from the dataset. Each annotation refers to a 33ms slice. On top, the track of an insect. In the middle, a bird flying upwards, and a drone doing a slow descending on the bottom figure. Color represents time.}
    \label{fig:example3d}
\end{figure}

To enable this study, we also present \dataset{}, a dataset of flying objects collected with an event camera (see Fig. \ref{fig:eyecatcher} for an overview of the annotations and Fig. \ref{fig:example3d} for examples of 3D event clouds of objects extracted from the dataset). To the best of our knowledge, this is the first neuromorphic dataset in the literature comprising manually annotated insects, birds and drones. Annotations include bounding boxes of flying objects, with temporal track identities.
In summary, the main contributions of this work are the following:
\begin{itemize}
    \item We propose \dataset{}, the first event-based flying objects dataset, including both insects, birds and drones.
    \item The dataset comprises spatio-temporal annotations, which can be leveraged for classification tasks by can be easily adopted for other tasks like detection and tracking.
    \item We characterize motion patterns of flying objects with arbitrarily low latency using point-based architectures.
\end{itemize}


    

\section{Related Works}

\subsection{Flying Object Recognition}
Flying object recognition can be broadly categorized into the detection of natural entities such as birds and insects, and artificial objects such as drones. The recognition of drones \cite{pawelczyk2020real, seidaliyeva2023advances, yousaf2022drone} has gained substantial interest due to security concerns, as they pose potential risks in unauthorized surveillance, airspace intrusion, and restricted zones. Traditional approaches to drone detection have relied on RGB cameras. Still, these methods are often challenged by factors such as the small size of drones, their rapid movements, and their ability to blend into complex backgrounds \cite{magrini2024neuromorphic}. To overcome these limitations, 
drone detection approaches have been studied also in the radio-frequency domain \cite{al2020drone}, with thermal imagery \cite{svanstrom2021real} or with acoustic sensors \cite{svanstrom2021real}.

Neuromorphic vision has emerged as a promising alternative, where event-based cameras, due to their ability to capture high-frequency temporal information, excel at detecting fast-moving drones and their propeller motion, which often presents a distinct signature in event streams \cite{sanket2021evpropnet}.
Recent works have also explored spiking neural networks for event-driven drone detection, leveraging their energy-efficient processing capabilities \cite{eldeborg2024low}.
Given the complementary nature of RGB and event-based vision, multimodal fusion techniques that combine both sensor types have demonstrated improved detection accuracy, as the two modalities compensate for each other’s limitations \cite{magrini2024neuromorphic}.

\begin{figure*}
    \includegraphics[width=\linewidth]{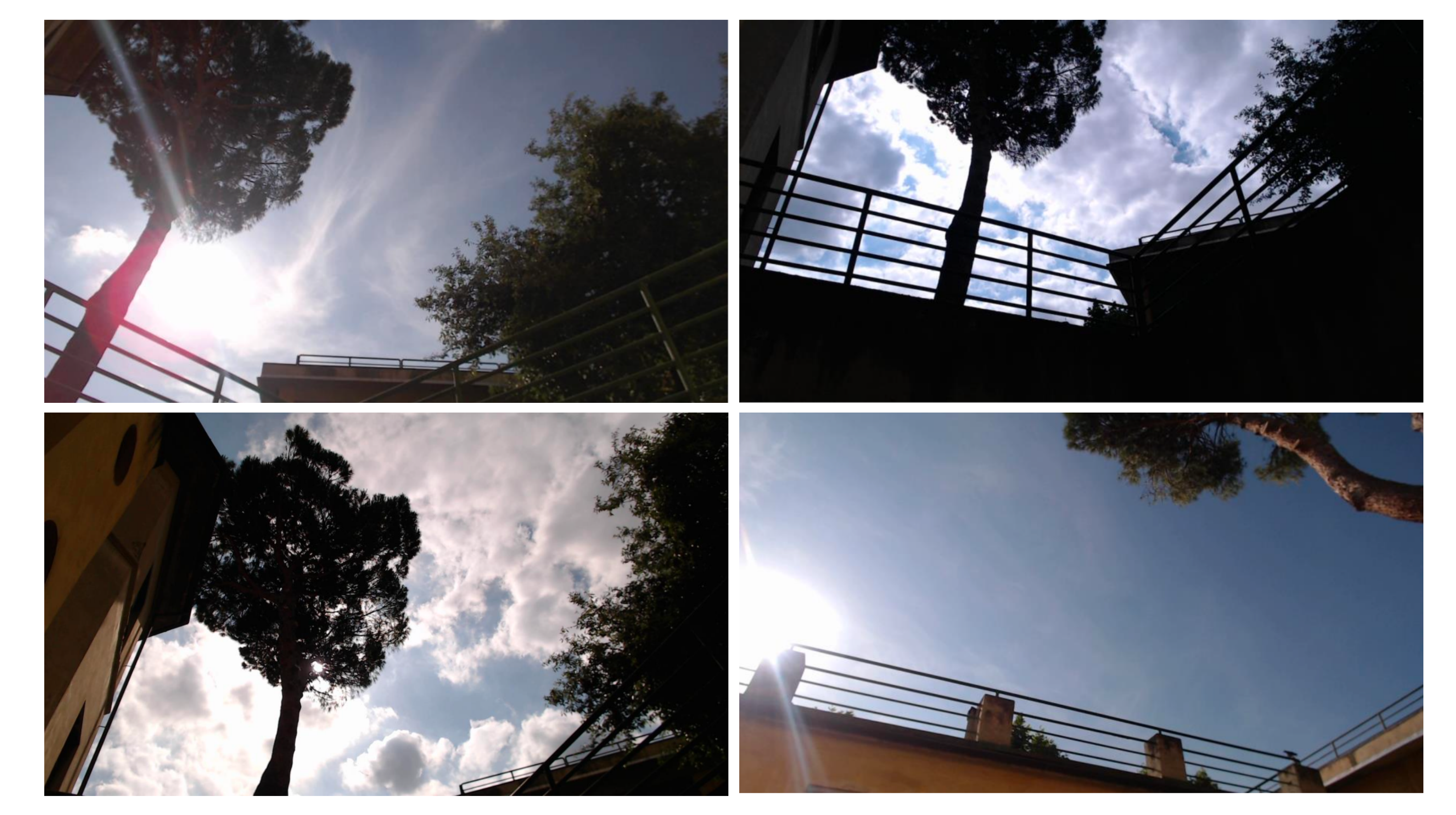}
    \caption{Some of the recording scenarios taken from an RGB equivalent camera during different times of the day. The presence of objects and obstacles in the scene leads to challenging occlusion scenarios.}
    \label{fig:rgbscene}
\end{figure*}

\subsection{Animal Monitoring}
The study of animal behavior and movement has traditionally relied on RGB-based monitoring techniques, where camera traps have been deployed to capture wildlife activity in natural habitats \cite{bubnicki2024camtrap, cultrera2023leveraging, hobbs2017improved, bjerge2022real}.
In particular, the detection of insects is a critical aspect of both ecological research and agricultural monitoring \cite{bjerge2023accurate, sittinger2024insect, naqvi2022camera, vo2024high, gardiner2024towards}. Deep learning approaches such as YOLO have been applied to track individual bees, allowing for detailed analysis of their movement and behavior \cite{ratnayake2021tracking}. These methods have been extended to more complex ecological interactions, including plant-insect interactions \cite{naqvi2022camera}, where insect activity around flowers is monitored to study pollination dynamics \cite{ratnayake2021towards}.
While RGB cameras have been the dominant sensing modality, event cameras have recently been introduced into animal monitoring applications due to their ability to capture high-speed motion and since they exhibit low consumption rates.
Event-based monitoring has demonstrated potential in tracking the flight patterns of birds and insects, which are often difficult to analyze with conventional frame-based approaches due to motion blur and frame-rate limitations \cite{pohle2024stereo, pohle355features, gebauer2024towards, sobue2017flying}.
Neuromorphic sensors have also been recently adopted for other types of animals including penguins \cite{hamann2024low} and mice \cite{hamann2024mousesis}.

\subsection{Event-Based Processing Methods}
Event cameras provide a fundamentally different way of capturing visual information compared to traditional frame-based sensors \cite{gallego2020event, chakravarthi2024recent}. Instead of recording images at fixed intervals, event cameras asynchronously detect changes in pixel intensity, leading to a sparse and temporally precise representation of motion. This unique sensing mechanism has inspired various processing techniques tailored to event data. In recent years, neuromorphic sensors have found applications in several fields of research including gesture recognition \cite{amir2017low}, video reconstruction \cite{zhang2022unifying}, face analysis \cite{becattini2024neuromorphic} and object detection \cite{gehrig2023recurrent}.

One of the most common approaches to process event streams is to accumulate events into frames, that can then be fed to models such as convolutional neural networks \cite{nguyen2019real,miao2019neuromorphic,ghosh2019spatiotemporal,cannici2020differentiable}.
In this work, we follow an alternative approach, by processing events as a point cloud.
By interpreting events as a 3D point cloud, each event is represented by its spatial coordinates (x, y) and timestamp t. This formulation allows for the direct application of point cloud processing techniques, which, unlike traditional images, capture continuous motion over time, making them particularly effective for fast-moving objects such as drones and insects.
Several works have treated event streams as point clouds.
Wang et al. \cite{wang2019space} applies a PointNet++ \cite{qi2017pointnet++} on an event rolling buffer for online gesture recognition, while the usage of PointNet \cite{qi2017pointnet} for extracting features from event cloud chunks in a masked autoencoder architecture has also been studied \cite{sun2025event}.
EventNet is a modified version of PointNet for real-time processing of event data \cite{sekikawa2019eventnet}, which has a PointNet backbone for event-based human pose estimation.
Spiking alternatives to PointNet have also been proposed \cite{ren2024spikepoint, ren2024spiking} and, recently, EventMamba has been proposed as a point-based version of Mamba \cite{ren2024rethinking}.
Other recent works have modeled event clouds with self-attention in event neighboring graphs and Gumbel subset sampling \cite{yang2019modeling}, or used VecKM \cite{yuan2024linear} to replace PointNet's dense local geometry encoding to improve its performance, as done in \cite{yuan2024learning} for event data.

\begin{table}[t]
  \centering
  \begin{tabular}{l c c c c}
    \hline
    \textbf{Statistics of \dataset{}} \\
    \hline
    Total duration & 67.34 minutes \\
    Total Object IDs & 3206 \\
    Total bounding boxes & 43869 \\
    Avg boxes per Object ID & 12.66 \\
    \hline
  \end{tabular}
  \caption{Statistics of the \dataset{} dataset.}
  \label{tab:dataset_statistics}
\end{table}

\section{The \dataset{} Dataset}

We propose \dataset{}, a novel dataset comprising more than one hour of event data depicting flying objects, recorded in the wild. In particular, the dataset is divided into 27 separate recordings, in different scenarios and angles, ensuring variability of recorded motion patterns of the flying objects. The annotations were made manually and covered 3 different types of objects, namely insects, birds and drones. Each class presents samples from various distances with the camera.
Recordings were manually annotated with bounding boxes at a $33ms$ granularity, for a total of 43.869 annotations.

Annotators labeled sequences of frames generated by accumulating all the events and adhering to the following protocol.
Each recording is analyzed frame-by-frame until a new object enters the frame. All subsequent frames where it appears are then labeled, assigning to it the same unique object identifier, and ignoring additional objects that may have appeared in the meantime.
After completing the annotation of the current object, the annotator returns to the first frame where the object appeared, and continues the frame-by-frame analysis until the next object appears.
A set of events that through subsequent frames develops a motion pattern that can be confidently attributed to that of a flying object is always annotated, regardless of the number of events that compose the resulting track.
Bounding boxes are defined around objects occupying the smallest possible area, to avoid overlaps with nearby objects or inclusion of noise or background events.
Only tracks that are at least 3 frames long ($99ms$) are retained as valid tracks.
Class categories (\texttt{insect}, \texttt{bird}, \texttt{drone}) are assigned to the final tracks. 

\begin{figure*}[t]
    \centering
    \includegraphics[width=.85\linewidth]{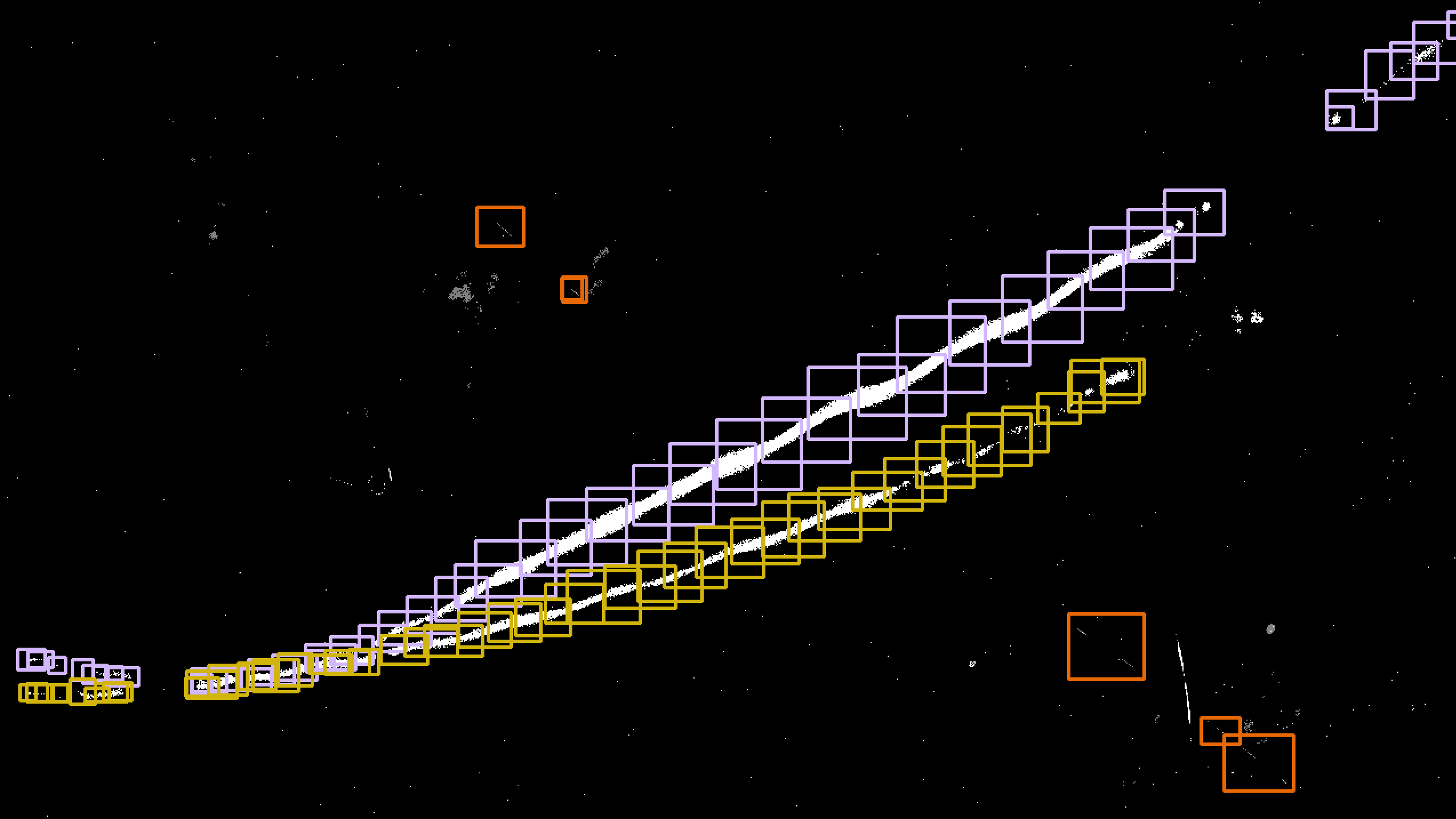}
    \caption{Samples from the \dataset{} dataset. Here we accumulate events for two seconds in the same frame, depicting bounding boxes annotated at 30Hz. Boxes are colored by track identity.}
    \label{fig:EV-FlyingFaunaExample}
\end{figure*}

All the recordings have been made using the Prophesee EVK2 camera, with a full-HD resolution of 1280x720. Fig. \ref{fig:rgbscene} shows the different recording environments used to capture the footage.
Tab.~\ref{tab:dataset_statistics} reports a summary of the frame-based annotations.
In addition, we also sample random spatio-temporal patches, non-overlapping with the annotated objects. We use these patches to provide a fixed set of negative examples for training and testing an object classifier that also takes into account non-objects.
Fig. \ref{fig:EV-FlyingFaunaExample} shows a sample extracted from the dataset. In the image, we accumulate events for two seconds, displaying also all the bounding box annotations present in such time interval. It can be noticed how tiny insects are clearly visible with an event camera. 
Instead, in Fig. \ref{fig:example3d} an example for each class (insect, bird, drone) is shown as a 3D point cloud.

\section{Problem Formulation}

Given a stream of events $E$, we assume we have access to an oracle capable of providing spatio-temporal coordinates of each recorded flying object, yielding a set of $N_E$ tracks $T(E)=\{\tau^1, ..., \tau^{N_E}\}$.
Each track $\tau^i$ in $T(E)$ is described by a sequence of events $\tau^i = \{t^i_j, x^i_j, y^i_j, p^i_j\} ~\forall i \in 1...N_{E}$. The number of events in $\tau^i$ can vary depending on the content of the event stream. We denote the number of events in track $\tau^i$ with $K^i$.

We then formulate the task of Flying Object Classification as the task of assigning a categorical label $\hat{y} \in \{\texttt{insect}, \texttt{bird}, \texttt{drone}, \texttt{background}\}$ to a given track.
We evaluate the predictions by computing the accuracy between the predicted categorical labels $\hat{y}$ and the ground truth labels $y$.

Furthermore, we establish two separate evaluation protocols, one for partial tracks and one for entire tracks. For the entire tracks, we process the whole set of events that composes it, regardless of its duration.
In the partial track evaluation, we split each track into chunks of fixed duration $\Delta t$ and treat each chunk as a separate example. This mimics a more realistic scenario, where a classification can be provided by observing a track only for a limited time span.



\begin{table*}
    \centering
    \renewcommand{\arraystretch}{1.2} 
    \setlength{\tabcolsep}{8pt} 
    \begin{tabular}{l|ccc|ccc}
    \toprule
        \multicolumn{1}{c}{} & \multicolumn{3}{c|}{\textbf{PointNet}} & \multicolumn{3}{c}{\textbf{PointNet++}} \\ \hline
        Sampling Method & 512 & 1024 & 2048 & 512 & 1024 & 2048 \\ \hline
        Random Sample      & 44.11  & 43.05  & 49.89  & \textbf{69.27}  & 67.71  & 69.35  \\
        Most Recent Sample & \textbf{46.57}  & \textbf{52.06}  & \textbf{55.27}  & 68.24  & 66.97  & 71.10  \\
        FPS                & 44.27  & 47.09  & 48.35  & 66.68  & \textbf{67.66}  & \textbf{72.06}  \\ 
        \bottomrule
    \end{tabular}
    \caption{Comparison of different sampling methods (Random, Most Recent, and FPS) for PointNet and PointNet++ on a 33ms frame split. The best result for each column is highlighted in bold.}
    \label{tab:sampling_comparison}
\end{table*}

\section{Proposed Model}
To address the flying objects identification tasks, we propose a simple and efficient point-based approach.
First, we leverage a point-based backbone model, namely PointNet~\cite{qi2017pointnet} and PointNet++~\cite{qi2017pointnet++}. Extending such methods to event clouds rather than point clouds is straightforward, as we treat each event as a 4-dimensional point, as done by previous works \cite{wang2019space, sun2025event, sekikawa2019eventnet, yang2019modeling}.
Such models provide point-wise features that are then aggregated into a single descriptor via a pooling operator, e.g. max pooling. This allows us to seamlessly process a variable number of event and obtain a learnable encoding of the whole cloud.

We hypothesise the usage of such network on realistic, real-time scenarios, where fast and lightweight models are preferred and we are usually bounded by computational and power constraints. 

In particular, given a detected bounding box - even using simple algorithmic methods, such as Metavision's spatter tracker\footnote{\url{https://docs.prophesee.ai/stable/samples/modules/analytics/tracking_spatter_py.html}} - we want to know whether the point cloud corresponding to the 2D coordinates of the frame represents a specific type of flying object. In this way, we can leverage the complete implicit spatio-temporal information of the point cloud to better classify the moving object based on its local movement, without the problem of quantizing the original information into event frames. 

To do so, we have to face some implicit challenges, in particular the best trade-off between speed and the used number of points. In our approach, we decided to sample a point cloud every 33ms, so to be aligned with the annotation frequency as well as to match the rate of most commercial RGB cameras. Once we have the 33ms corresponding point cloud, we sample it into a smaller number of points to be more representative of the local movement of the object and to reduce the possible impact of excessive background noise. In particular, we propose 3 different sampling strategies, namely Farthest Point Sampling (FPS), Random Sampling and Most Recent Sampling. 
While the first focuses on sampling the points with the best overall coverage of the original point cloud, it is also more prone to capturing unwanted noise. The second strategy instead relies on its simplicity, and can withstand moderately noisy clouds. Lastly, we propose Most Recent Sampling as the last N events along the temporal axis of the point cloud, or the most recent ones. In this case, the sampling gives more weight to the more recent point cloud structure.

\section{Experiments}

\subsection{Flying Object Classification}

To test the goodness of the proposed approach and the novel dataset, we first experiment on the classification task of single tracks. In particular, we tested on both full track classification and single chunk classification. The latter means that we split the whole track into slices of fixed time, and we perform classification on the single split using only its points. 
While the full track approach gives us a good estimate of the long-term classification capability, the single chunk gives us more insights on the resilience of such approach in critical scenarios where the object is present only for a small fraction of time at maximum. 

The pipeline is the following: we slice each track extracted from all the videos into fixed sequences of 33ms, and treat each of them separately. We pass them to either a PointNet or PointNet++ feature extractor, and use the said feature on a 3-layer MLP with a final linear layer for classification. 

In the case of the full track classification, we perform an additional majority voting on the classification results of the single slices.

\subsection{Single Chunk Classification}

Classification accuracies for the single chuck scenario are reported in Tab.~\ref{tab:sampling_comparison}, where we vary the number of points to be sampled in $\{512, 1024, 2048\}$. As we can see, PointNet++ demonstrates its superior performance in all scenarios and sampling strategies, proving to be less susceptible to the sampling choice. Still, both Pointnet and PointNet++ show a clear advantage when more points are being sampled from the single temporal chunk. 

At the same time, while PointNet performs better with the Most Recent sampling, PointNet++ seems to show better results when using the Farthest Point Sampling, probably given the multi-scale sampling nature also present in PointNet++.

The confusion matrices for different sampling points and strategies for the PointNet++ based modelv are shown in Fig.~\ref{img:conf_mats}, where 0 represents the class \texttt{background}, 1 the class \texttt{bird}, 2 the class \texttt{insect} and 4 the class \texttt{drone}. From the results it is clearly visible how th e model is able in most situation to correctly discern the background from actual flying objects, while is not always able to correctly separate insects from the birds and drones in many scenarios.

\subsection{Whole Track Classification}
We also test the whole track classification for PointNet++, and the results are reported in Tab.~\ref{tab:pointnetpp_results}. Here, we can see how the model performs better on different numbers of sampling points based on the sampling strategy. Overall, the best full-track classification accuracy is reached by the Most Recent Sampling on 1024 sampled points. 

Overall, we can say that 1024 represents a good trade-off between computational expense and overall performance, with the proposed Most Recent Sampling strategy performing well in all models and scenarios. 

Finally, it is worth noticing how a good accuracy in single chunk accuracy does not strictly correlate with long-term accuracy, showing the challenging scenario for both temporally local and global classification of the presented dataset.

\newcommand{\confwidth}{.30\textwidth}

\begin{figure*}[t]
\centering
    \includegraphics[width=\confwidth]{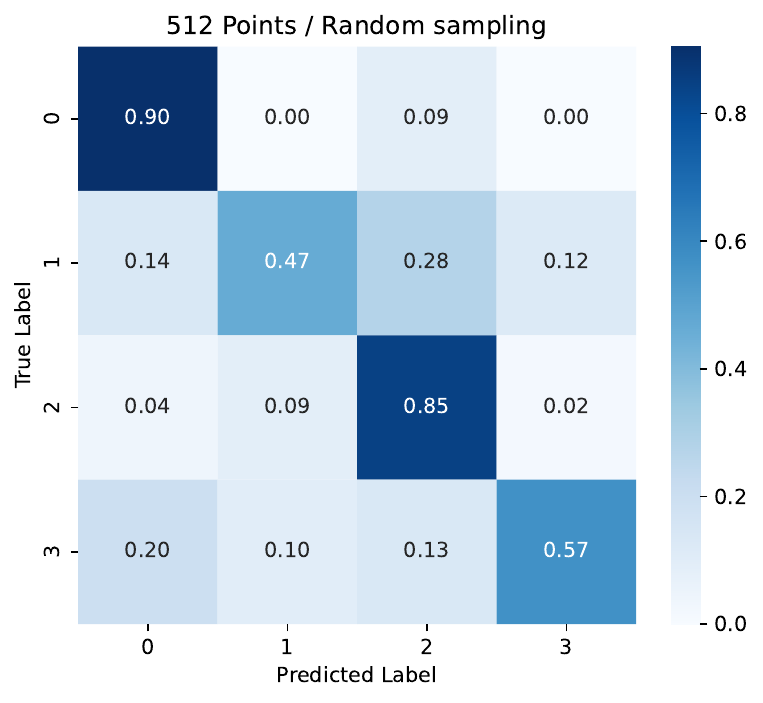}
    \includegraphics[width=\confwidth]{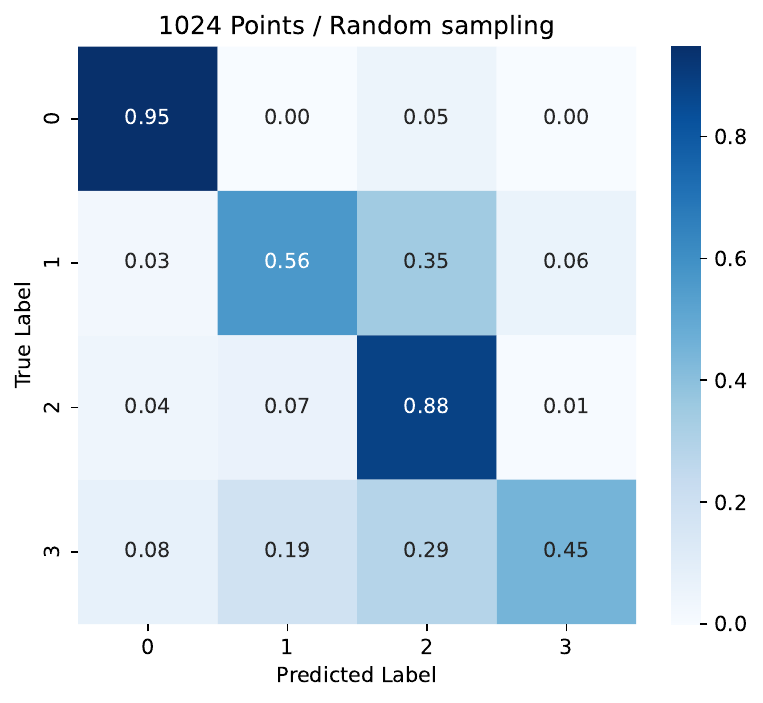}
    \includegraphics[width=\confwidth]{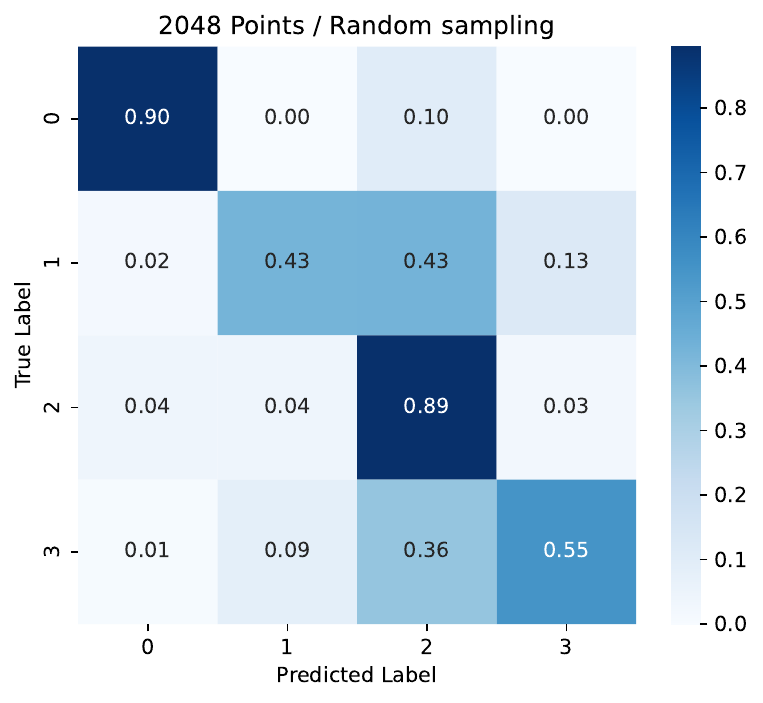}
    \\
    \includegraphics[width=\confwidth]{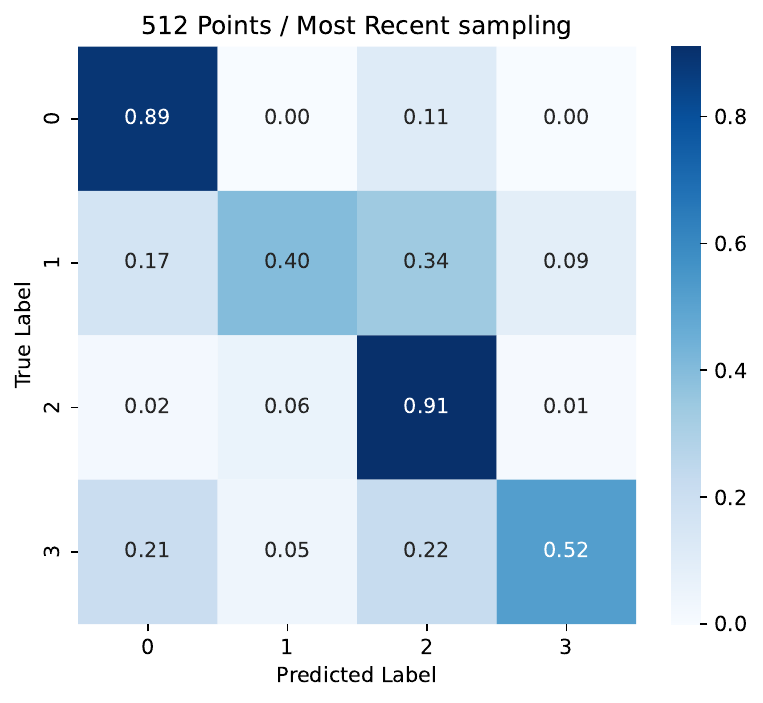}
    \includegraphics[width=\confwidth]{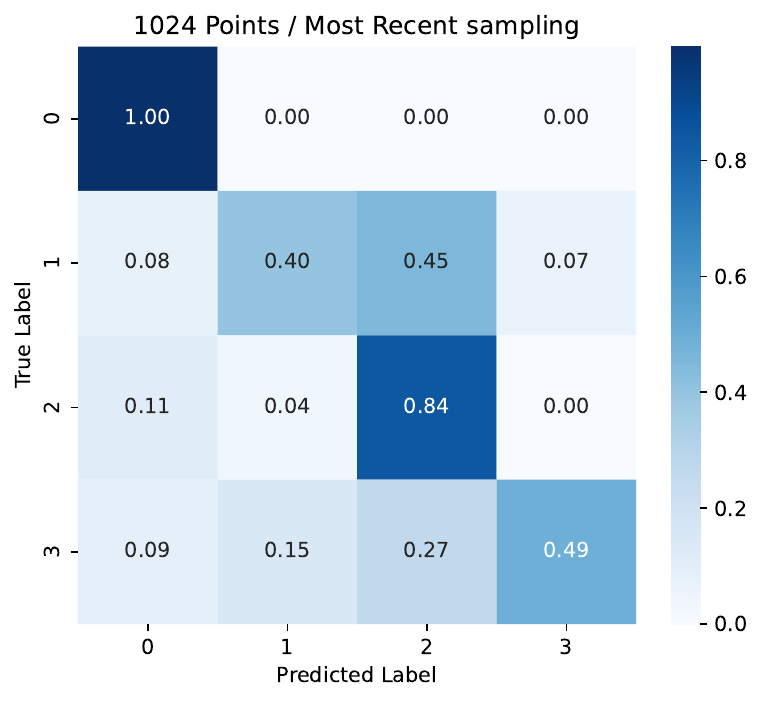}
    \includegraphics[width=\confwidth]{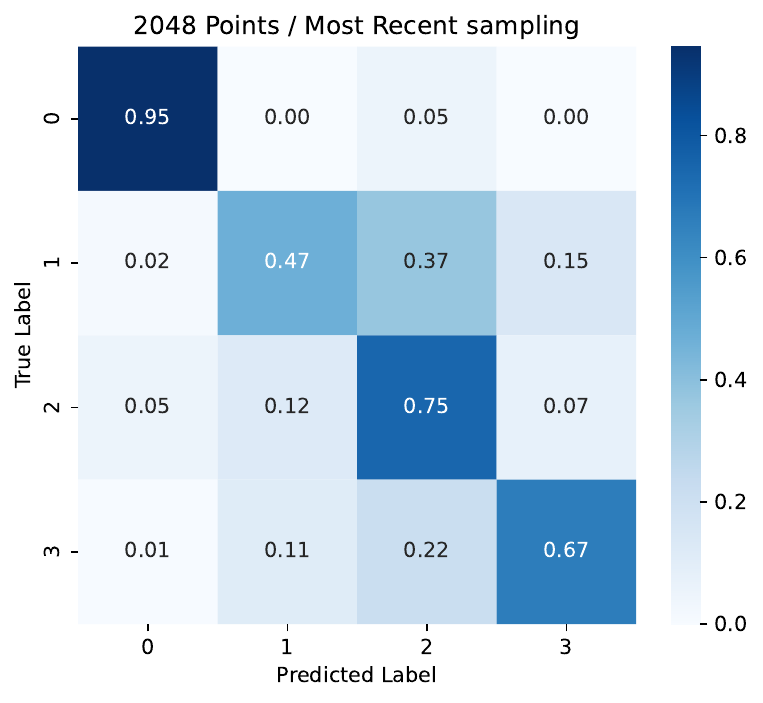}
    \\
    \includegraphics[width=\confwidth]{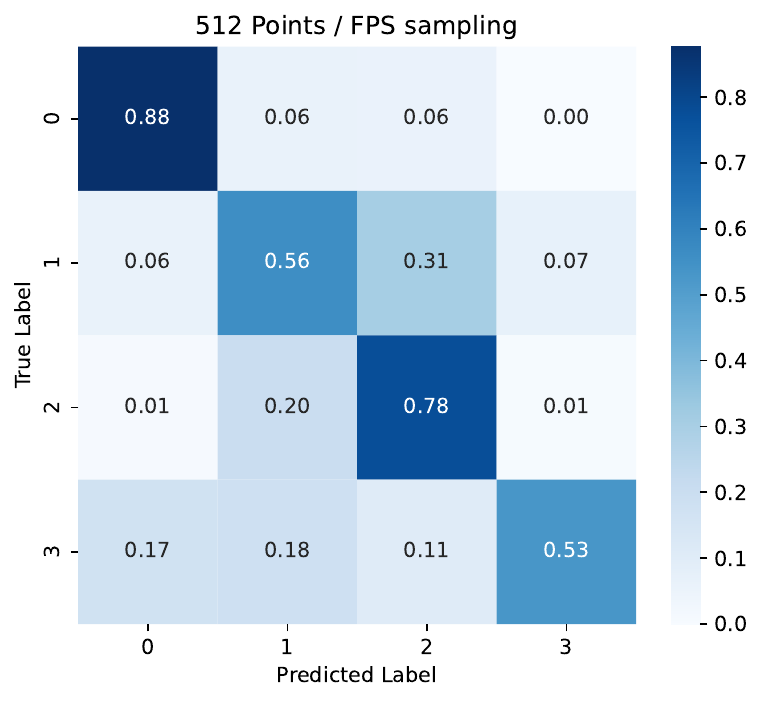}
    \includegraphics[width=\confwidth]{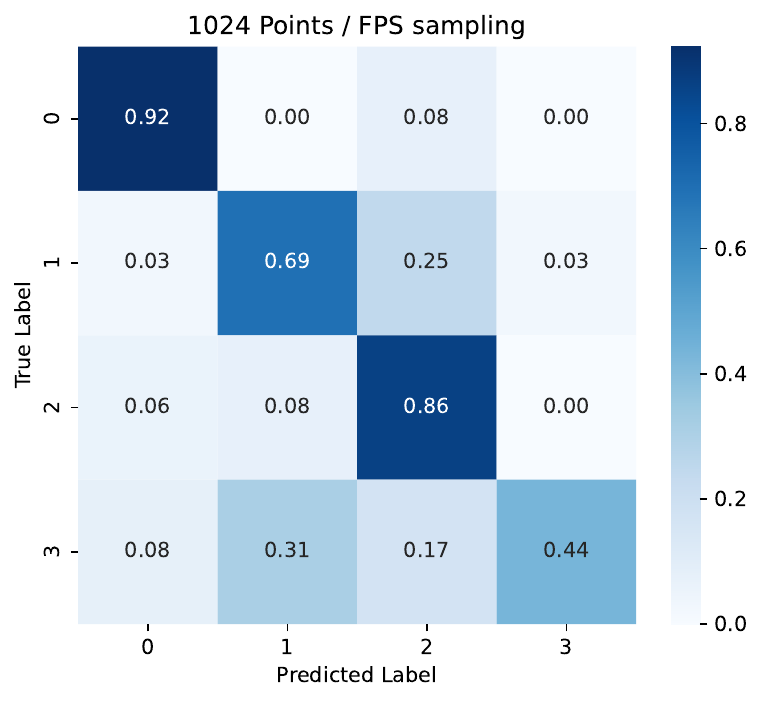}
    \includegraphics[width=\confwidth]{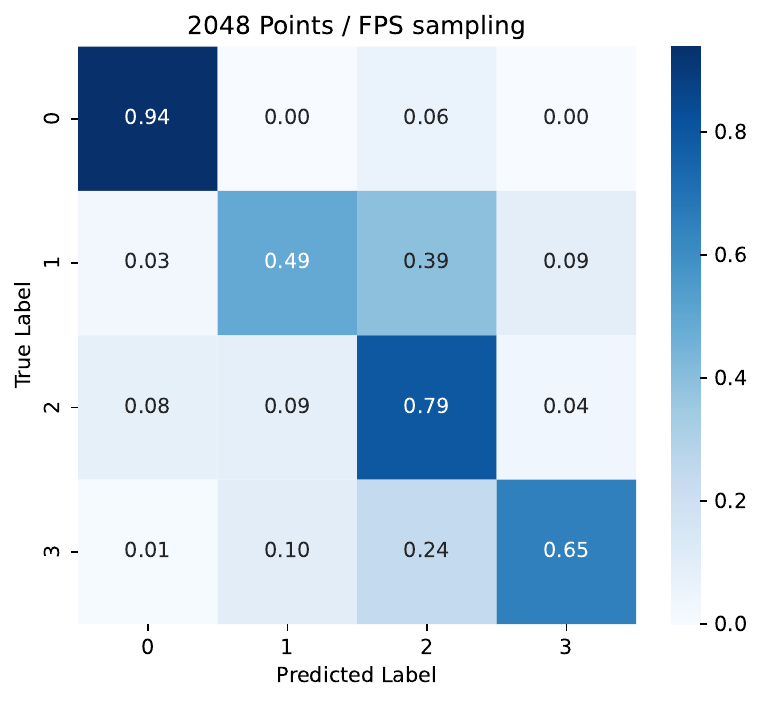}
    \caption{Confusion matrices for different numbers of events and sampling strategies using PointNet++.}
    \label{img:conf_mats}
\end{figure*}

\begin{table}
    \centering
    \renewcommand{\arraystretch}{1.2} 
    \setlength{\tabcolsep}{10pt} 
    \begin{tabular}{l|ccc}
        \toprule
        \textbf{Sampling Method} & \textbf{512} & \textbf{1024} & \textbf{2048} \\ 
        \midrule
        Random Sample      & \textbf{90.53}  & 91.65  & 85.52  \\
        Most Recent Sample & 76.25  & \textbf{92.02}  & 78.75  \\
        FPS                & 55.10  & 78.85  & \textbf{88.68}  \\ 
        \bottomrule
    \end{tabular}
    \caption{Full track classification performance of different sampling methods (Random, Most Recent, and FPS) for PointNet++ on a 33ms frame split in the Majority Voting setting. The best result for each column is highlighted in bold.}
    \label{tab:pointnetpp_results}
\end{table}

\section{Ethical Considerations and Impact}
The development of event-based vision for aerial object detection carries important ethical implications, particularly regarding privacy, security, and ecological impact. The potential for continuous and autonomous monitoring of flying objects, including drones and wildlife, necessitates responsible data collection practices to prevent misuse in surveillance applications. Ethical deployment must ensure compliance with legal frameworks governing airspace monitoring, wildlife research, and data protection. Additionally, while this technology enhances conservation efforts by enabling non-intrusive monitoring of birds and insects, care must be taken to minimize unintended ecological disturbances. The long-term impact on biodiversity studies and environmental policies could be significant, offering a novel tool for sustainable observation and analysis.
Furthermore, the increasing prevalence of drones raises additional ethical and regulatory challenges. While drones are valuable for applications such as search and rescue, environmental monitoring, and infrastructure inspection, their potential for unauthorized surveillance and airspace violations must be addressed. The ability of event-based vision to detect and classify drones in real time can enhance security and air traffic management, mitigating risks associated with unauthorized or hazardous drone activity. However, such systems must be implemented with safeguards to prevent misuse, such as unauthorized surveillance or infringement on personal freedoms.

\section{Conclusions}
In this work, we introduced EV-Flying, the first event-based dataset for recognizing and classifying flying objects, including birds, insects, and drones. By leveraging the advantages of neuromorphic vision, such as high temporal resolution, low latency, and robustness to motion blur, we demonstrated the effectiveness of event-based sensors for aerial object detection. Our study employed a point-based approach inspired by PointNet, allowing for efficient processing of asynchronous event streams while preserving fine-grained motion information. Experimental results highlighted the effectiveness of event-based vision in distinguishing different flying entities.

\section*{Acknowledgments}
This work was partially funded by "Collaborative Explainable neuro-symbolic AI for Decision Support Assistant, CAI4DSA, CUP B13C23005640006".
This work was partially supported by the Piano per lo Sviluppo della Ricerca (PSR 2023) of the University of Siena - project FEATHER: Forecasting and Estimation of Actions and Trajectories for Human-robot intERactions.

{
    \small
    \bibliographystyle{ieeenat_fullname}
    \bibliography{main}
}


\end{document}